\documentclass[letterpaper, 10 pt, conference]{ieeeconf}  

\IEEEoverridecommandlockouts                              

\newcommand{\RNum}[1]{\uppercase\expandafter{\romannumeral #1\relax}}

\usepackage{graphicx,wrapfig,fullpage,amsmath,hhline,epsfig,verbatim,url,amssymb,multicol,multirow,cite}
\usepackage{times,color,soul} 
\usepackage[left=0.75in,top=0.70in,right=0.75in,bottom=0.80in]{geometry}
\usepackage{amsbsy,latexsym,mathrsfs,mathtools}

\usepackage{mathptmx} 
\DeclareMathAlphabet{\mathcal}{OMS}{cmsy}{m}{n}
\usepackage{bm}
\usepackage{float}
\usepackage{color,soul}
\usepackage{dsfont}
\usepackage{comment}
\usepackage{psfrag}   
\usepackage{url}
\usepackage{hyperref}
\usepackage{setspace}
\usepackage[table]{xcolor}
\usepackage{paralist}
\usepackage{booktabs}
\usepackage{balance}

\usepackage{algorithm} 
\usepackage{algorithmic}  
\usepackage[algo2e]{algorithm2e} 
\newtheorem{thm}{Theorem}

\usepackage{soul}

\title{\LARGE \bf
	 Time-Varying ALIP Model and Robust Foot-Placement Control for Underactuated Bipedal Robot Walking on a Swaying Rigid Surface
}

\author{Yuan Gao$^1$, Yukai Gong$^{2}$, Victor Paredes$^3$, Ayonga Hereid$^3$, Yan Gu$^4$
\thanks{$^{1}$Y. Gao is with the College of Engineering, University of Massachusetts Lowell, Lowell, MA 01854, USA.
{yuan\_gao@student.uml.edu.}}%
\thanks{$^{2}$Y. Gong is with the Robotics Department, University of Michigan, Ann Arbor, MI 48105, USA. ykgong@umiche.edu}%
\thanks{$^{3}$V. Paredes and A. Hereid are with the Department of Mechanical and Aerospace Engineering, the Ohio State University, Columbus, OH 43210, USA.
{paredescauna.1@buckeyemail.osu.edu, hereid.1@osu.edu.}}%
\thanks{$^{4}$Y. Gu is with the School of Mechanical Engineering,
Purdue University, West Lafayette, IN 47907, USA.
{yangu@purdue.edu.}}%
}

\begin{document}

\maketitle
\thispagestyle{plain}
\pagestyle{plain}
\begin{abstract}

Controller design for bipedal walking on dynamic rigid surfaces (DRSes), which are rigid surfaces moving in the inertial frame (e.g., ships and airplanes), remains largely uninvestigated.
This paper introduces a hierarchical control approach that achieves stable underactuated bipedal robot walking on a horizontally oscillating DRS.
The highest layer of our approach is a real-time motion planner that generates desired global behaviors (i.e., the center of mass trajectories and footstep locations) by stabilizing a reduced-order robot model.
One key novelty of this layer is the derivation of the reduced-order model by analytically extending the angular momentum based linear inverted pendulum (ALIP) model from stationary to horizontally moving surfaces.
The other novelty is the development of a discrete-time foot-placement controller that exponentially stabilizes the hybrid, linear, time-varying ALIP model.
The middle layer of the proposed approach is a walking pattern generator that translates the desired global behaviors into the robot's full-body reference trajectories for all directly actuated degrees of freedom.
The lowest layer is an input-output linearizing controller that exponentially tracks those full-body reference trajectories based on the full-order, hybrid, nonlinear robot dynamics.
Simulations of planar underactuated bipedal walking on a swaying DRS confirm that the proposed framework ensures the walking stability under difference DRS motions and gait types.

\end{abstract}

\section{Introduction}
	Bipedal robots can aid in various critical real-world applications such as search and rescue, emergency response, and warehouse management.
	Those applications may demand robots to navigate on nonstationary walking platforms, such as shipboard firefighting, inspection, and maintenance.
    Enabling stable legged locomotion on a nonstationary rigid platform, which we call a dynamic rigid surface (DRS)~\cite{9847283}, is a fundamentally challenging control problem due to the high complexity of the robot dynamics that is nonlinear, hybrid, and time varying~\cite{9108552}.
    To that end, the objective of this study is to derive and validate a hierarchical control approach that enables stable bipedal underactuated walking on a rigid swaying surface (e.g., a vessel's deck).

    \subsection{Related Work}
    
    Various control approaches have been created to realize provably stable bipedal robot walking on stationary rigid surfaces, among which the most widely studied one is the hybrid zero dynamics (HZD) method~\cite{grizzle2001asymptotically}.
    The HZD approach stabilizes bipedal walking by explicitly treating the full-order, hybrid, nonlinear robot dynamics.
    For underactuated robots (e.g., bipeds with point feet), the HZD method exploits input-output linearization to transform the nonlinear robot dynamics associated with the directly actuated degrees of freedom (DOFs) into a linear time-invariant system, which is then stabilized based on the well-studied linear system theory.
    Due to the use of input-output linearization, internal dynamics exist, and its solutions (e.g., periodic orbits) are typically unstable for walking robots.
    The HZD method constructs a reduced-order zero dynamics manifold that agrees with the overall hybrid dynamics and searches for stable periodic orbits on that manifold.
    \begin{figure}[t]
    \centering
    \includegraphics[width=1\linewidth]{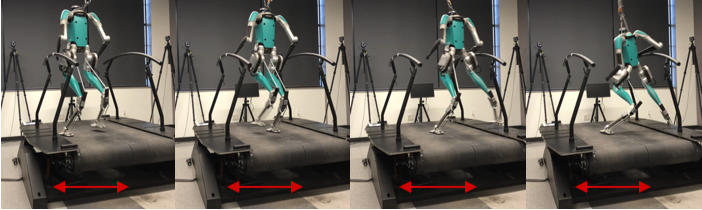}
    \vspace{-0.3in}
    \caption{The default controller of the Digit humanoid robot seems to fail to guarantee stable walking on a DRS that sways at a frequency of $0.5$ Hz and a magnitude of $5$ cm.}
    \label{Fig:digit_fail.png}
    \vspace{-0.3in}
\end{figure}
    
    Due to the high dimensionality and strong nonlinearity of a full-order robot model, real-time generation of stable desired trajectories based on the full-order model can be computationally prohibitive for achieving robust bipedal walking on stationary uneven terrains. 
    To that end, 
    researchers have integrated reduced-order model based planning with full-order model based control.
    X. Xiong et al. developed a hybrid linear inverted pendulum (LIP) model to approximate the hybrid walking dynamics of an underactuated bipedal robot~\cite{xiong20223, xiong2021global,dai2022bipedal}. 
    Y. Gong et al. proposed a new variant of the LIP model that uses the angular momentum about the contact point, instead of the linear velocity of CoM, as a state variable~\cite{gong2021zero,gong2021one}, which is called the ``ALIP''.
    V. Paredes et al. introduced a LIP template model to generate a stepping controller with an adaptive learning regulator to ensure stable walking on a bipedal humanoid robot~\cite{paredes2022resolved}.
    Yet, due to the time-varying movement of the surface-foot contact point/region, the dynamic model of bipedal walking on a DRS is explicitly time-varying~\cite{iqbal2021modeling,9108552}, which is fundamentally different the typical time-invariant robot dynamics during static-surface locomotion.

    Recently, the control problem of stabilizing legged locomotion on a DRS has been initially studied.
    To provably stabilizes quadrupedal walking on a vertically moving DRS,
    A. Iqbal et al. introduced a nonlinear control approach that explicitly handles the time-varying DRS accelerations and the hybrid, nonlinear robot dynamics during quadrupedal walking~\cite{9108552}.
    To enable physically feasible and computationally efficient planning for legged locomotion on a vertically moving surface, the classical continuous-time LIP model has been analytically extended, resulting in a time-varying, homogeneous LIP model~\cite{iqbaloptimization,iqbal2022drs}.  
    Still, the walking robot considered is fully actuated, and thus the inherent instability associated with underactuated walking does not exist. 
    Also, the surface is assumed to move only in the vertical direction in these studies.
    In fact, our recent experiment validation of the proprietary controller of the Digit humanoid robot (developed by Agility Robotics) seems to indicate that horizontally moving surfaces might be substantially more challenging for bipedal robots to handle (see Fig.~\ref{Fig:digit_fail.png}).
    

    \subsection{Contribution}
    This study introduces a hierarchical control approach that achieves stable underactuated bipedal walking on a swaying rigid surface by explicitly treating the robot's hybrid, time-varying dynamics and simultaneously exploiting the complementary advantages of full-order and reduced-order models for walking stabilization.
    The specific contributions are:
    \begin{itemize}
        \item [a)] Analytically extending the ALIP model from stationary surfaces to a horizontally swaying DRS, resulting in a hybrid time-varying ALIP model.
        \item [b)] Synthesizing a discrete-time footstep controller that exponentially stabilizes the hybrid, time-varying ALIP model, and formulating an optimization problem to find the desired CoM trajectories and the stabilizing footstep locations.
        \item [c)] Developing a three-layer control approach that ensures the stability for the hybrid, time-varying, nonlinear unactuated robot dynamics by stabilizing the proposed ALIP model and by mitigating the model inaccuracy of the ALIP model with proper full-order trajectory design and tracking control.
        \item [d)] Demonstrating the proposed approach enables a full-order robot to stably walk on a DRS under different surface motions and gait types.
    \end{itemize}

    This paper is structured as follows.
    Section~\ref{Section:ALIP model} introduces the derivation of a hybrid time-varying ALIP model for bipedal walking on a swaying DRS.
    Section~\ref{Section ALIP planner} proposes a discrete-time footstep controller that exponentially stabilizes the hybrid ALIP model for DRS walking, and presents the formulation of the higher-layer planner of the proposed approach that produces the desired global trajectories.
    Section~\ref{Section-Implementing ALIP} explains the translation of the desired global trajectories into the full-body references trajectories for the directly controlled variables.
    Section~\ref{Sec: I/O linearizing control} presents the lower-layer controller that exponentially tracks the desired full-body trajectories.
    Section~\ref{Section-Simulation} reports simulation validation results.
    Section~\ref{section-conclusions} provides the concluding remarks. 

\section{Time-Varying Angular Momentum Based Linear Inverted Pendulum (ALIP) Model}
\label{Section:ALIP model}

This section introduces the derivation of the proposed ALIP model that captures the essential robot dynamics associated with bipedal walking on a horizontally swaying DRS.
The ALIP model serves as the basis of the higher-layer planner introduced in Sec. III.

Different from the classical LIP model~\cite{kajita20013d} that takes a robot's center of mass (CoM) position and velocity as its state, we choose to use the CoM position and the angular momentum about the surface-foot contact point as the state, resulting in an ALIP model~\cite{gong2021zero}.
Compared with the classical LIP model, such a state choice allows the ALIP model to be more accurate in representing the true robot dynamics in the presence of velocities jumps at foot landings, large peak motor torques, or aggressive swing leg motions~\cite{gong2021zero}.

\subsection{Angular Momentum about Foot-Surface Contact Point}

This subsection introduces the mathematical expression of a bipedal robot's angular momentum about the foot-surface contact point. 
For simplicity, this study only considers bipeds with point feet. 

\subsubsection{Continuous swing phase}
During a swing phase, one foot of the biped contacts the walking surface, and the other moves in the air.
Let $S$ denote the contact point attached to the walking surface.
For a DRS, the point $S$ moves in the world frame.
Let $A$ denote a stationary point in the world frame that instantaneously coincides with $S$ at the given time.

During a single-support phase, the three-dimensional (3-D) vector of angular momentum about the point $S$, denoted as $\mathbf{L}_S$, has the following relation with the angular momentum about the point $A$, denoted as $\mathbf{L}_A$:
\vspace{-0.05in}
\begin{equation}
\small
    \mathbf{L}_S = \mathbf{L}_{A}
    +
    \mathbf{p}_{SA}\times (m\mathbf{v}_{CoM}).
    \label{Eq: LA and LS}
\vspace{-0.05in}
\end{equation}
Here the vector $\mathbf{p}_{SA}$ is the position of point $A$ relative to the point $S$, expressed in the world frame.
Note that $\mathbf{p}_{SA} = \mathbf{0}$ because the point A instantaneously coincides with the point S at the given time. 
The scalar constant $m$ is the total mass of the robot.
The vector $\mathbf{v}_{CoM}$ is the absolute CoM velocity with respect to (w.r.t.) the world frame.

We will use the relation in Eq.~\eqref{Eq: LA and LS} to derive the dynamics of $\mathbf{L}_S$ for legged locomotion on a DRS (Sec. II-B).

\subsubsection{Discrete foot-switching event}
At the end of the swing phase, the swing foot touches the walking surface.
Without loss of generality, we assume that the support foot begins to swing just after the swing foot touchdown. 

Across a foot landing event, the position of the contact point jumps.
Let $k$ ($k \in \{1,2,... \}$)
indicate the $k^{th}$ foot landing event.
Let $\mathbf{L}_{k}$ be the angular momentum about the $k^{th}$ contact point on the walking surface.
Let $\mathbf{p}_{(k+1)\rightarrow k}$ denote the position vector pointing from the $(k+1)^{th}$ to the $k^{th}$ contact point.
Let $(\cdot)^-$ and $(\cdot)^+$ respectively represent the values of $(\cdot)$ just before and after the $k^{th}$ foot-landing instant.

Just before the $k^{th}$ foot-landing instant, the angular momentum about the new contact point, $\mathbf{L}_{(k+1)}^-$, can be related to the previous one, $\mathbf{L}_{k}^-$, as follows:
\vspace{-0.05in}
\begin{equation}
\small
\label{equ:AM transfer}
    \mathbf{L}^-_{k+1} = \mathbf{L}^-_{k}+\mathbf{p}_{(k+1)\rightarrow k}\times (m\mathbf{v}^-_{CoM}).
\vspace{-0.05in}
\end{equation}

Across the $k^{th}$ foot-landing event, the surface-foot impact at the new contact point generates zero impulse torque about the $(k+1)^{th}$ contact point.
Thus, the angular momentum about the $(k+1)^{th}$ contact point, $\mathbf{L}_{k+1}$, does not change across a foot-landing impact; that is, $\mathbf{L}^+_{k+1}=\mathbf{L}^-_{k+1}$.
Accordingly, $\mathbf{L}^+_{k+1}$ can be expressed as:
\vspace{-0.05in}
\begin{equation}
\small
\label{equ:AM transfer2}
    \mathbf{L}^+_{k+1} = \mathbf{L}^-_{k}+\mathbf{p}_{(k+1)\rightarrow k}\times m\mathbf{v}^-_{CoM}.
    \vspace{-0.05in}
\end{equation}

\noindent \textbf{Remark 1 (Invariance to foot switching events for 2-D cases):}
From Eq.~\eqref{equ:AM transfer2}, we notice that when a planar robot walks on a flat, horizontal surface with zero vertical CoM velocity, $\mathbf{L}^+_{k+1} = \mathbf{L}^-_{k}$ holds; that is, the robot's angular momentum about the contact point is invariant to the foot switching event~\cite{gong2021one}.

\begin{figure}[!t]
    \centering
    \includegraphics[width=1\linewidth]{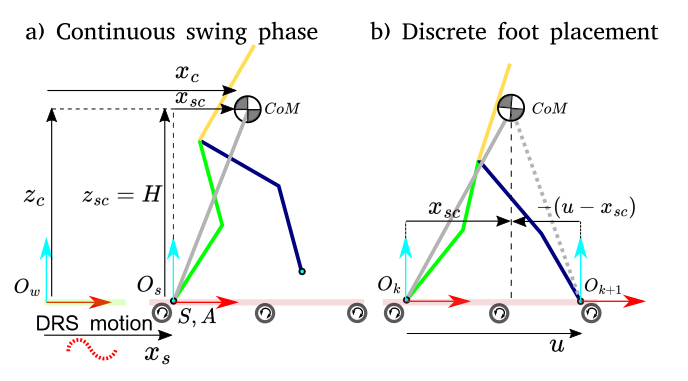}
    \vspace{-0.3in}
    \caption{Illustration of the proposed hybrid AIP model a) during a continuous swing phase and b) at a discrete foot-landing event.}
    \label{Fig:LIP}
    \vspace{-0.2in}
\end{figure}

\subsection{2-D ALIP Model during a Continuous Swing Phase}

This subsection introduces the proposed ALIP model that captures the essential continuous-phase robot dynamics associated with 2-D walking on a DRS (see Fig.~\ref{Fig:LIP}-a).

As explained in subsection A, we choose to use the angular momentum about the contact point as a state variable of the proposed ALIP, in addition to the CoM position. 
To derive the needed dynamic model with these state variables, we first derive the accurate dynamic model in 3-D and then provide the approximate ALIP model in 2-D.

\subsubsection{Deriving $\dot{\mathbf{L}}_S$}

Recall that the point A is a static point in the world frame that instantaneously coincides with the contact point S (which is attached to the DRS) at the given time and that the 3-D angular momentum vectors $\mathbf{L}_A$ and $\mathbf{L}_S$ are related through Eq.~\eqref{Eq: LA and LS}.
Taking the time derivative of both sides of  Eq.~\eqref{Eq: LA and LS} yields
\vspace{-0.05in}
\begin{equation}
\small
\label{equ:dLs}
    \dot{\mathbf{L}}_S = \dot{\mathbf{L}}_A+ \dot{\mathbf{p}}_{SA}\times m\mathbf{v}_{CoM}+\mathbf{p}_{SA}\times m\mathbf{\dot{v}}_{CoM}.
    \vspace{-0.05in}
\end{equation}
Since $\mathbf{p}_{SA} = \mathbf{0}$ and $\dot{\mathbf{p}}_{SA} =- \dot{\mathbf{p}}_{S}$, Eq.~\eqref{equ:dLs} becomes
\vspace{-0.05in}
\begin{equation}
\small
\label{equ:dLs2}
    \dot{\mathbf{L}}_S = \dot{\mathbf{L}}_A- \dot{\mathbf{p}}_{S}\times m\mathbf{v}_{CoM}.
    \vspace{-0.05in}
\end{equation}

As the time derivative of the angular momentum $\mathbf{L}_A$ equals the sum of the external moments about point $A$, we have
\vspace{-0.05in}
\begin{equation}
\small
\label{Eq: dLA}
\dot{\mathbf{L}}_A = \mathbf{p}_{AC} \times (m \mathbf{g}) + \boldsymbol{\tau}_A,
\vspace{-0.05in}
\end{equation}
where $\mathbf{p}_{AC}$ is the position of the CoM relative to point A, $\mathbf{g}$ is the gravitational acceleration, and $\boldsymbol{\tau}_A$ is the external torque that is applied to the contact point.

For a robot with point feet, $\boldsymbol{\tau}_A=\mathbf{0}$.
Also, $\mathbf{p}_{AC}=\mathbf{p}_{SC}$ where $\mathbf{p}_{SC}$ is the CoM position relative to point S.
Thus, Eq.~\eqref{Eq: dLA} can be rewritten as:
\vspace{-0.05in}
\begin{equation}
\small
\label{Eq: dLA2}
\dot{\mathbf{L}}_A = \mathbf{p}_{SC} \times (m \mathbf{g}).
\vspace{-0.05in}
\end{equation}

Combining Eqs.~\eqref{equ:dLs2} and~\eqref{Eq: dLA2}, we obtain the 3-D dynamic model in terms of $\mathbf{L}_S$: 
\vspace{-0.05in}
\begin{equation}
\small
\label{Eq: dLS3}
\dot{\mathbf{L}}_S = 
\mathbf{p}_{SC} \times (m \mathbf{g})
-
\dot{\mathbf{p}}_{S}\times (m\mathbf{v}_{CoM}).
\vspace{-0.05in}
\end{equation}

\subsubsection{Deriving $\dot{\mathbf{p}}_{SC}$}

The relative CoM velocity $\dot{\mathbf{p}}_{SC}$ can be expressed as: 
\vspace{-0.05in}
\begin{equation}
\small
    \dot{\mathbf{p}}_{SC}
    =
    \mathbf{v}_{CoM}
    -
    \dot{\mathbf{p}}_{S},
    \vspace{-0.05in}
\end{equation}
where the expression of $\mathbf{v}_{CoM}$ can be obtained through the following relationship between $\mathbf{L}_S$ and the robot's angular momentum about the CoM, denoted as $\mathbf{L}_{CoM}$:
\vspace{-0.05in}
\begin{equation}
\small
    \mathbf{L}_S = \mathbf{L}_{CoM}
    +
    \mathbf{p}_{SC}\times (m\mathbf{v}_{CoM}).
    \label{Eq: LS and LCOM}
    \vspace{-0.05in}
\end{equation}

\subsubsection{2-D ALIP model}

Similar to the classical LIP model and the ALIP model for stationary-surface locomotion, 
we assume that the CoM height above the DRS remains constant.
Also, for simplicity, we assume that the DRS only moves horizontally, which is realistic for real-world DRSes such as the deck of a vessel moving in regular sea waves~\cite{gahlinger2000cabin,benjamin1954stability}.
Thus, the linear velocity of the DRS at the support point S, $\dot{\mathbf{p}}_{S}$, and the CoM velocity, $\mathbf{v}_{CoM}$, are parallel for 2-D robots.
Accordingly, $\dot{\mathbf{p}}_{S}\times (m\mathbf{v}_{CoM})=\mathbf{0}$ holds for 2-D robots, with which Eq.~\eqref{Eq: dLS3} becomes:
\vspace{-0.05in}
\begin{equation}
\small
\label{Eq: dLS4}
\dot{\mathbf{L}}_S = 
\mathbf{p}_{SC} \times (m \mathbf{g}).
\vspace{-0.05in}
\end{equation}

By approximating a biped as an inverted pendulum (i.e., a point mass atop a massless leg)~\cite{kajita20013d} [cite],
the angular momentum about the CoM becomes zero; i.e., 
$\mathbf{L}_{CoM}=\mathbf{0}$.
Thus, Eq.~\eqref{Eq: LS and LCOM} becomes:
\vspace{-0.05in}
\begin{equation}
\small
    \mathbf{L}_S = 
    \mathbf{p}_{SC}\times (m\mathbf{v}_{CoM}).
    \label{Eq: LS and LCOM-2}
    \vspace{-0.05in}
\end{equation}

Since this study considers 2-D bipeds, we can express the angular momentum $\mathbf{L}_S$ as a scalar variable, which is denoted as $L_S$.
Also, under the assumption that the CoM height is constant, we only need to consider the 1-D horizontal movement along the $x$-direction as shown in Fig. 1. 

Let $(x_{SC},z_{SC})$ be the coordinates of the position vector $\mathbf{p}_{SC}$.
Note that $\dot{z}_{SC}=0$ under the assumption of the constant CoM height, and let $H$ denote the constant CoM height (i.e., $z_{SC}=H$).
Let $\dot{x}_{SC}$ and $\dot{x}_{S}$ be the forward CoM velocity relative to the point S (i.e., the $x$-component of $\dot{\mathbf{p}}_{SC}$) and the horizontal surface velocity (i.e., the $x$-component of $\dot{\mathbf{p}}_{S}$), respectively. 
The 3-D vector equations~\eqref{Eq: LS and LCOM}-\eqref{Eq: LS and LCOM-2} can be written in the scalar form as
\vspace{-0.05in}
\begin{equation}
\small
    \dot{L}_{S} = mg x_{SC}
    \quad \mbox{and} \quad
    \dot{{x}}_{SC} = \frac{L_S}{m H}-\dot{x}_{S}.
    \vspace{-0.05in}
\end{equation}

Accordingly, with the state definition $\mathbf{x}:=[x_{SC},L_S]^T$, the state-space representation of the 2-D ALIP model during a continuous swing phase is given by:
\vspace{-0.05in}
\begin{equation}
\small
\label{equ:ALIPDRS}
\underbrace{
    \begin{bmatrix}
        \dot{x}_{SC}
        \\
        \dot{L}_S
    \end{bmatrix}
    }_{\dot{\mathbf{x}}}
    =
    \underbrace{
    \begin{bmatrix}
        0 & \frac{1}{mH}
        \\
        mg & 0
    \end{bmatrix}
    }_{=:\mathbf{A}}
    \underbrace{
    \begin{bmatrix}
        {x}_{SC}
        \\
        {L}_S
    \end{bmatrix}
    }_{\mathbf{x}}
    -
    \underbrace{
    \begin{bmatrix}
        \dot{x}_{S}(t)
        \\
        0
    \end{bmatrix}
    }_{=:\mathbf{f}(t)}.
    \vspace{-0.05in}
\end{equation}




\noindent \textbf{Remark 2 (Time-varying ALIP model):}
Note that the proposed continuous-phase ALIP model in Eq.~\eqref{equ:ALIPDRS} is explicitly time-varying due to the time-varying forcing term $\mathbf{f}(t)$.
Such a time-varying property differs from the standard time-invariant LIP model described in~\cite{kajita20013d}.
The explicit time dependence of the forcing term $\mathbf{f}(t)$ is caused by the time-varying velocity of the support point S, $\dot{x}_S(t)$.
As any point on a rigid surface has the same linear velocity, $\dot{x}_S(t)$ represents the linear surface motion.
In this study, we assume that the surface motion $\dot{x}_S(t)$ is continuous and differentiable, and is periodic with  ${x}_S(t)={x}_S(t+T)$ for any $t>0$ and some real, positive constant $T$.

\subsection{2-D ALIP Model at A Foot Landing}

As explained in subsection A, a biped's two feet instantaneously switch their roles at a foot landing event, causing a sudden jump in the contact point position ${x}_S$ and the relative CoM position $x_{SC}$ (Fig.~\ref{Fig:LIP}-b).
Let $u$ denote the new support point position of the ALIP relative to the previous one. Then, $u=x_S^+ - x_S^-$.
Let $\Delta x_{SC}$ be the change in $x_{SC}$ across a foot landing event.
Then, $\Delta x_{SC}:=x_{SC}^+ - x_{SC}^-=x_S^- - x_S^+ = - u$.

As highlighted in Remark 1, $L_S$ does not jump across a foot landing due to its invariance to the foot landing events~\cite{gong2021one}.
Thus, the change of $L_S$ upon a foot landing satisfies: $\Delta L_S := L_S^+ - L_S^-=0$.

Let $\Delta \mathbf{x}$ be the difference between the pre- and post-switching state.
Then, the jump/impact map of $\Delta \mathbf{x}$ is compactly given by:
\vspace{-0.05in}
\begin{equation}
\small
\label{equ:discrete ALIP}
\begin{bmatrix}
    \Delta x_{SC}
    \\
    \Delta L_S
\end{bmatrix}
=
\begin{bmatrix}
    -u
    \\
    0
\end{bmatrix}.
\vspace{-0.05in}
\end{equation}
\noindent \textbf{Remark 3 (Fixed-time switching for ALIP):}
We set the $k^{th}$ $(k=1,2...)$ switching event for the proposed ALIP model to occur at periodic, fixed time instants $t=\tau_k^-$ with $\tau_k$ a real scalar and $\tau_{k+1}=\tau_{k}+T$.
Setting the switching events as fixed-time instead of state-triggered helps simplify the stability analysis for the overall hybrid ALIP model.
Also, adopting a fixed-time switching is realistic since we only use the ALIP model to plan the desired motion, although the switching of the actual motion is not fixed-time and is affected by early or late foot landings~\cite{gu2018exponential}.

\begin{figure}[!t]
    \centering
    \includegraphics[width=1\linewidth]{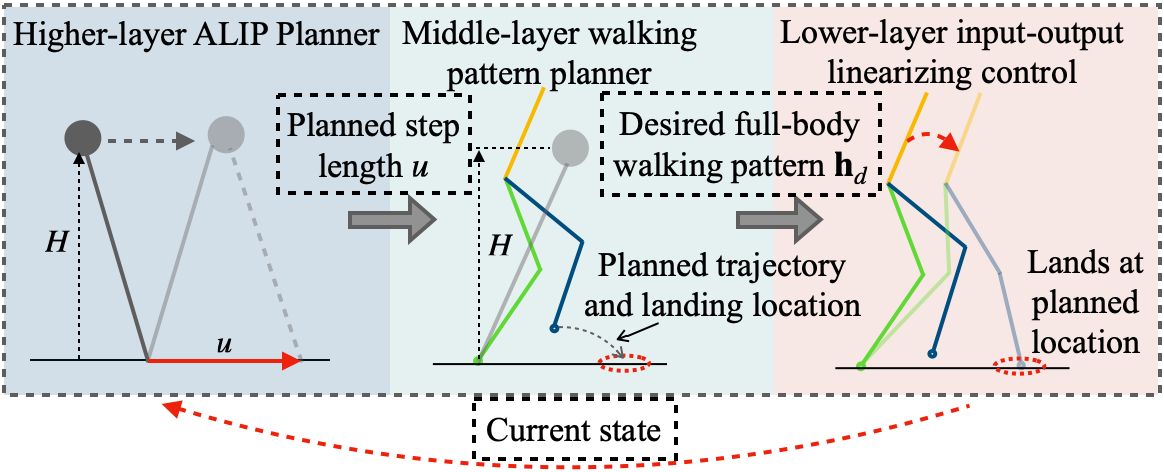}
    \vspace{-0.25 in}
    \caption{Overview of the proposed hierarchical control approach.
    The ALIP-based planner generates the desired foot landing locations based on the current state of the full-order model.
    The middle-layer walking pattern generator adjusts the desired swing foot trajectories based on the full-order model and the higher-layer planning result.
    The input-output linearizing controller reliably tracks the full-order reference trajectories.}
    \label{Fig: frame illustartion}
    \vspace{-0.2in}
\end{figure}

\section{Higher-Layer ALIP-based Planner}
\label{Section ALIP planner}

This section presents the higher-layer planner of the proposed hierarchical control approach.
This layer generates the desired CoM trajectory and foot placement for the lower-layer controller to reliably track.

As illustrated in Fig. 2, the higher-layer ALIP-based planner is part of the proposed three-layer control approach that aims to enable stable underactuated DRS walking on a full-order robot.
The key challenge in achieving this control objective is the complexity of the unactuated robot dynamics that is hybrid, time-varying, and nonlinear.

To achieve this control objective, the rationale of the overall proposed control approach is to exploit the higher-layer planner to stabilize the unactuated dynamics based on its approximate model (i.e., the proposed ALIP) while using the lower-layer controller to stabilize the directly actuated degrees of freedom based on their accurate full-order model.
To handle the unactuated dynamics of the robot, we use the ALIP model to approximately represent the unactuated dynamics, and the higher layer directly stabilizes the reduced-order ALIP model, thus indirectly ensuring the stability for the unactuated dynamics of the actual robot.
Meanwhile, the middle and lower layers serve two purposes towards the overall control objective.
First, to mitigate the model discrepancies between the real robot and the ALIP model, the middle layer generates full-body reference trajectories that are compatible with the ALIP model (e.g., by enforcing model simplifying assumptions underlying the ALIP such as the constant CoM height and zero angular momentum about the CoM).
Second, the middle layer translates the desired footstep locations supplied by the higher layer into reference swing foot trajectories that the lower layer controller reliably tracks, thus ensuring the actual robot faithfully executes the planned foot placement for ensuring the stability for the unactuated dynamics of the actual full-order robot.

The proposed ALIP-based planner begins with the derivation of a discrete-time foot placement controller based on the hybrid, time-varying ALIP model, as explained next.

\subsection{Discrete-Time Foot Placement Control}
The continuous-phase ALIP model in Eq.~\eqref{equ:ALIPDRS} is unstable and has no control input.
This subsection introduces the derivation of a control strategy at the foot switching instant that stabilizes the overall hybrid ALIP model.
The control command is the footstep position of the ALIP, which is also referred to as swing-foot-landing control~\cite{xiong20223}.

Inspired by~\cite{xiong20223}, we formulate the footstep controller as:
\vspace{-0.05in}
\begin{equation}
\small
\label{equ:step length}
    u = u^*+\mathbf{K}(\mathbf{x}^- - \mathbf{x}^*),
    \vspace{-0.05in}
\end{equation}
where $u^*$ is the desired footstep location of the ALIP.
The vector $\mathbf{K}$ is the feedback gain to be designed with $\mathbf{K}\in\mathbb{R}^{1\times 2}$.
Here, the vector $\mathbf{x}^-$ is the actual pre-impact state of the ALIP.
The vector $\mathbf{x}^*$ is the desired pre-impact state of the ALIP.

From Eqs.~\eqref{equ:discrete ALIP} and \eqref{equ:step length}, we obtain the closed-loop discrete-time dynamics at a foot landing event:
\vspace{-0.05in}
\begin{equation}
\small
\label{equ:feedback ALIP}
\underbrace{
    \begin{bmatrix}
        \Delta x_{SC}
        \\
        \Delta L_S
    \end{bmatrix}
    }_{\Delta \mathbf{x}}
    =
    \underbrace{
    \begin{bmatrix}
        -\mathbf{K}
        \\
        \mathbf{0}_{1\times 2}
    \end{bmatrix}
    }_{=:\mathbf{B}}
    \underbrace{
    \begin{bmatrix}
        x_{SC}^-
        \\
        L_S^-
    \end{bmatrix}
    }_{\mathbf{x}^-}
    +
    \underbrace{
    \begin{bmatrix}
        \mathbf{K}\mathbf{x}^*-u^*
        \\
        0
    \end{bmatrix}
    }_{=:\mathbf{G}}, \quad t = \tau_k^- ~ (k=1,2,...).
        \vspace{-0.05in}
\end{equation}

\subsection{Stability Conditions}
The hybrid ALIP system described by Eqs.~\eqref{equ:ALIPDRS} and~\eqref{equ:feedback ALIP} has the following expression:
    \vspace{-0.05in}
\begin{equation}
\small
\label{equ:hybrid system}
\begin{cases}
    \begin{aligned}
        \mathbf{\dot{x}} &= \mathbf{A}\mathbf{x}^- - \mathbf{f}(t), &t\neq\tau_k^- ~(k=1,2,...);
        \\
        \Delta \mathbf{x} &= \mathbf{B}\mathbf{x}^-+\mathbf{G}, &t = \tau_k^- ~(k=1,2,...).
    \end{aligned}
\end{cases}
    \vspace{-0.05in}
\end{equation}
According to~\cite{bainov1993impulsive}, the stability condition for the periodic solution of such a linear time-varying nonhomogeneous hybrid system is as follows:

\begin{thm}
    If all eigenvalues of the monodromy matrix associated with the homogeneous equation of the system in Eq.\eqref{equ:hybrid system} are less than one in modulus, then the origin of the homogeneous equation is exponentially stable.
    Moreover, the $T$-periodic solution $\boldsymbol{\psi}(t)$ (with $\boldsymbol{\psi}(t+T)=\boldsymbol{\psi}(t)$) of the hybrid non-homogeneous system is exponentially stable.
\end{thm}


\subsection{Exponential Stabilization of Periodic ALIP Walking through Optimization}

We formulate an optimization problem to find a periodic solution $\boldsymbol{\psi}(t)$ (with period $T$) of the hybrid ALIP model in Eq.~\eqref{equ:hybrid system} that respects the physical constraints of the actual robot, as well as to find the control gain $\mathbf{K}$ that exponentially stabilizes $\boldsymbol{\psi}(t)$.

\subsubsection{Optimization variables} 
The optimization variables include $\mathbf{K}$, $u^*$, and $\mathbf{x}^*$ as they are the parameters that define the footstep controller in Eq.~\eqref{equ:step length}.

\subsubsection{Cost function}
We choose to minimize the norm of $\mathbf{K}$ to help prevent the controller from overreacting to small values of $(\mathbf{x}^- - \mathbf{x}^*)$.
Thus, the cost function is $J = \mathbf{K}\mathbf{K}^T$. 

\subsubsection{Constraints}
The constraints considered comprise the stability and the feasibility constraints:
\begin{itemize}
    \item [(a)] The stability constraint ensures that the solution of the ALIP model exponentially converges to the planned trajectory by enforcing the stability condition in Theorem 1.
    Let $\mu_i$ be the $i^{th}$ eigenvalue of the monodromy matrix associated with the homogeneous portion of the hybrid ALIP model in Eq.~\eqref{equ:hybrid system}.
    To ensure stability, Theorem 1 requires $|\mu_i|<1$ for any $i$.
    \item [(b)] The feasibility constraint ensures that the full-order model can achieve the desired trajectory without violating its kinematic limits.
    Let $u_{min}$ and $u_{max}$ be the minimum and maximum feasible values of the step length $u$.
    Let $\mathbf{x}_{min}$ and $\mathbf{x}_{max}$ be feasible lower and upper bounds of the state variable $\mathbf{x}$.
    Then, a physically feasible solution should satisfy: $u_{min}\leq u^* \leq u_{max}$ and $\mathbf{x}_{min}\leq \mathbf{x}^* \leq \mathbf{x}_{max}$.
\end{itemize}

\subsubsection{User-defined gait parameters}
To plan the desired walking motion of a legged robot, users typically specify the key features of the desired robot behaviors. 
Also, we assume that the surface motion is known (i.e., either measured or estimated), which is realistic for real-world moving platforms (e.g., vessels) that are equipped with motion monitoring systems~\cite{yoon2008development}. 
Thus, we pre-specify the following parameters and variables as input to the optimization problem: CoM height $H$, gait period $T$, DRS motion profile $x_S(t)$, and total mass of the robot $m$.

\subsubsection{Optimization formulation}
To summarize, the optimization problem can be formulated as follows:
    \vspace{-0.05in}
\begin{equation}\label{equ:opt}
\small
        \begin{aligned}
        & \text{Cost function}: && J = \mathbf{K}\mathbf{K}^T
        \\
        & \text{s.t.} 
        & &  u_{min}\leq u^* \leq u_{max} &~\text{(C-1)}
        \\
        & & &  \mathbf{x}_{min}\leq \mathbf{x}^* \leq \mathbf{x}_{max} &~\text{(C-2)}
         \\
        & & & |\mu_i|<1 &~\text{(C-3)}
        \end{aligned}
            \vspace{-0.05in}
\end{equation}

\begin{figure}[!t]
    \centering    \includegraphics[width=0.4\linewidth]{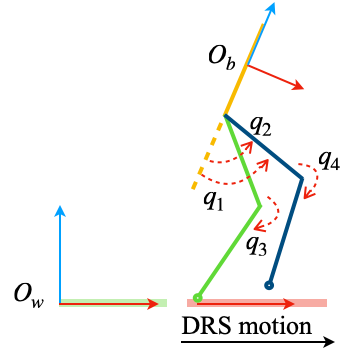}
    \vspace{-0.15in}
    \caption{Illustration of a full-order biped that walks on a swaying DRS.}
    \label{Fig:2d robot}
    \vspace{-0.15in}
\end{figure}

\vspace{-0.07in}
\section{Middle-Layer Walking Pattern Planner}

\label{Section-Implementing ALIP}

This section presents the middle-layer planner of the proposed control approach. This layer generates the full-body reference motions compatible with both the ALIP model and the desired global behaviors produced by the higher-layer planner. 
These full-body trajectories are tracked by the lower-layer controller explained in Sec. V.

\subsection{Full-Body Control Variable Selection}

Let $\mathbf{q}$ be the vector of the generalized coordinates of a full-order planar robot.
Typically, $\mathbf{q}$ is defined as $\mathbf{q}:=[p^b_x,p^b_y,\theta^b,q_1,...,q_n]$,
where the scalar variables $p^b_x$, $p^b_y$ and $\theta^b$ are the pose of the robot's base in the world frame and $q_i$ ($i=1,...,n$) is the angle of the $i^{th}$ revolute joint of this robot.
We assume the robot has point feet with all revolute joints powered (except ankles).

The number of the DOFs of a planar robot with point feet can be computed as
    \vspace{-0.05in}
\begin{equation*}
\small
    \text{DOF} = n+3-2=n+1,
        \vspace{-0.05in}
\end{equation*}
where $``3"$ is the DOF of the planar floating base and $``2"$ is the number of holonomic constraints associated with a planar point contact between the support foot and the surface.
As all joints except the point feet are powered, the number of actuators is $n$.
Because $\text{DOF}>n$, the planar walking robot is underactuated (with one degree of underactuation).

Let us consider a planar bipedal robot with four revolute joints (i.e., $n=4$) as shown in Fig.~\ref{Fig:2d robot}.
Let $\mathbf{h}_c$ denote the control variables:
    \vspace{-0.05in}
\begin{equation}
\small
\label{equ:hc}
    \mathbf{h}_c(\mathbf{q})=[z_{CoM}(\mathbf{q}),\theta^b,x_{sw}(\mathbf{q}),z_{sw}(\mathbf{q})]^T,
        \vspace{-0.05in}
\end{equation}
where $z_{CoM}$ is the CoM height, $\theta_b$ is the trunk orienattion, and
$x_{sw}$ and $z_{sw}$ are the swing foot position in $x$- and $z$-directions relative to the support foot.

We choose to directly control the CoM height $z_{CoM}$ to increase the accuracy of the ALIP.
Specifically, the CoM height of the actual robot should remain constant and match the ALIP model.
Also, we choose to control the trunk orientation $\theta_b$ to ensure an upright trunk posture and an approximately zero angular momentum about the CoM.
Finally, by controlling the swing foot position ($x_{sw}$,$z_{sw}$), the full-order robot can reliably execute the desired footstep locations supplied by the higher-layer planner.

\subsection{Full-Body Trajectory Generation}

Let $\mathbf{h}_d$ denote the vector of the desired trajectories for the control variables $\mathbf{h}_c$.
We define $\mathbf{h}_d$ as:
    \vspace{-0.05in}
\begin{equation}
\small
\label{equ: hd}
    \mathbf{h}_d = [\phi_1,\phi_2,\phi_3,\phi_4]^T,
        \vspace{-0.05in}
\end{equation}
where the scalar functions $\phi_1, ..., \phi_4$ represent the desired trajectories for the CoM height, trunk orientation, forward swing foot position, and vertical swing foot position.

\subsubsection{Trajectory parameterization}
We parameterize the desired trajectory $\phi_i$ using B\'ezier polynomials~\cite{westervelt2007feedback,gao2019dscc}.
To encode the B\'ezier polynomials, we  use a time-based phase variable $s$ that represents how long a step has progressed.
Let $T_{k}$ denote the actual switching instant at the end of the $(k)^{th}$ actual walking step of the full-order robot.
Then $s:=\frac{t-T_{k}}{T}$.
Thus, $s$ equals $0$ and $1$ at the beginning and the end of a planned step, respectively.

The B\'ezier polynomial $\phi_i$ is expressed as:
    \vspace{-0.05in}
\begin{equation*}
\small
    \label{Bezier polynomial}
	{\phi}_i(s):=\sum_{m=0}^{M}{\alpha}_{i,m} \frac{M!}{m!(M-m)!}s^m(1-s)^{M-m},
     \vspace{-0.05in}
\end{equation*}
where $M$ is the order of the B\'ezier polynomial and ${\alpha}_{i,m}$ is the $m^{th}$ coefficient.
One convenient property of the B\'ezier polynomials is $\phi_i(0)=\alpha_{i,0}$ and $\phi_i(1)=\alpha_{i,M}$, which relates the value of the coefficients to the value of the polynomial.
The desired trajectory can be adjusted by tuning coefficients.

\subsubsection{Trajectory and Phase Variable Update}
The higher-level planner constantly updates the B\'ezier coefficients of $\phi_3(s)$ based on the robot's actual horizontal footstep location. 
Let $(\cdot)_t$ denote the value of a variable at time step $t$.
The specific update procedure at time step $t$ is summarized in Algorithm 1:

 \begin{algorithm}
 \label{alg:ALIP update}
 \SetAlgoLined

  \While{True }{
   At time step $t$, obtain the current generalized coordinates $\mathbf{q}_t$ of the full-order robot.

   \If{
     s=0
   }
   {
   Assign the current swing foot position in the $x$-direction to $\alpha_{3,0}$: 
  
   ~~$\alpha_{3,0}$ $\leftarrow$ $x_{sw}(\mathbf{q}_t)$.
    }
    {

    }
    \eIf{
     $0\leq s \leq1$
   }{
     Convert $\mathbf{q}_t$ into the current ALIP state:

     ~~$\mathbf{x}_t\leftarrow\mathbf{x}(\mathbf{q}_t)$.

     Compute the pre-impact state of ALIP model $\mathbf{x}^-_t$ based on current state $\mathbf{x}_t$ by integrating the continuous-phase ALIP dynamics.
    

     Compute $u_t$ using $\mathbf{x}^-_t$ and based on Eq.~\eqref{equ:step length}.
  
     Assign the planned swing-foot landing location $u_t$ to $\alpha_{3,M}$:
    
     ~~$\alpha_{3,M} \leftarrow u_t$.

    Assign $\alpha_{3,1}$,..., $\alpha_{3,M-1}$ as the linear interpolation between $\alpha_{3,0}$ and $\alpha_{3,M}$.
   }
   {
     No update is applied to $\alpha_{3,0}$,..., $\alpha_{3,M}$.
   }
  }
  \caption{Pseudocode for updating the B\'ezier polynomial coefficients at time step $t$ based on the ALIP planner result}
 \end{algorithm}

\section{Lower-Layer Input-Output Linearizing Control based on Hybrid Full-Order Model}

\label{Sec: I/O linearizing control}

This section explains the feedback controller that we use to command the directly controlled variables of the actual robot to follow the desired full-body motion.

As inspired by the HZD approach~\cite{grizzle2001asymptotically}, we develop a tracking controller based on the full-order, hybrid, nonlinear robot dynamics model and the exact linearization of the nonlinear map between the joint torques and the directly commanded tracking error.
Also, similar to our recent work on the provable stabilization of DRS walking~\cite{9108552}, we explicitly handle the time-varying nature of the robot dynamics through modeling and controller design.

\subsection{Full-Order Dynamic Model}

\subsubsection{Full-order continuous-phase dynamics}

During the continuous phase of bipedal walking, a robot with point feet has point contact with the ground.
Let $\mathbf{p}_F(\mathbf{q})$ denote the position of the robot's support foot w.r.t. the world frame.
Assuming the support foot does not slip on the DRS (i.e., $\mathbf{p}_F(\mathbf{q})=\mathbf{p}_S(t)$),
the holonomic constraint associated with the surface-foot contact can be expressed as
    \vspace{-0.05in}
\begin{equation}
\small
    \label{equ:holo constraint} \mathbf{J}_F\mathbf{\ddot{q}}+\mathbf{\dot{J}}_F\mathbf{\dot{q}} = \mathbf{\ddot{p}}_S(t),
        \vspace{-0.05in}
\end{equation}
where $\mathbf{J}_F:=\frac{\partial \mathbf{p}_F}{\partial \mathbf{q}}(\mathbf{q})$.

With Lagrange's method, the continuous-phase dynamics of a bipedal robot that walks on a DRS can be expressed as:
    \vspace{-0.05in}
\begin{equation}
\small
\label{equ:full order dynamics}
    \mathbf{M}(\mathbf{q})\ddot{\mathbf{q}}+\mathbf{c}(\mathbf{q},\mathbf{\dot{q}})=\mathbf{B}\boldsymbol{\tau}_u+\mathbf{J}_F^T\mathbf{f}_F,
        \vspace{-0.05in}
\end{equation}
where $\mathbf{M}$ is the inertia matrix,
$\mathbf{c}$ is the sum of the gravitational, Coriolis, and centrifugal terms, $\mathbf{B}$ is the constant input-selection matrix,
$\boldsymbol{\tau}_u$ is a vector of joint torques,
and $\mathbf{f}_F$ is the ground reaction force induced by the interaction between the contact foot and the moving surface.
From Eqs.~\eqref{equ:holo constraint} and~\eqref{equ:full order dynamics}, we obtain
    \vspace{-0.05in}
\begin{equation}
\small
    \label{equ:full order dynamics with holo}
    \mathbf{M}(\mathbf{q})\ddot{\mathbf{q}}+\mathbf{\bar{c}}(t,\mathbf{q},\mathbf{\dot{q}})=\mathbf{\bar{B}}\boldsymbol{\tau}_u,
        \vspace{-0.05in}
\end{equation}
where
$\mathbf{\bar{c}}:= \mathbf{c}-\mathbf{{J}}^T_F(\mathbf{J}_F\mathbf{M}^{-1}\mathbf{{J}}^T_F)^{-1}(\mathbf{J}_F\mathbf{M}^{-1}\mathbf{c}-\mathbf{\dot{J}}_F\mathbf{\dot{q}}+\mathbf{\ddot{p}}_S (t))$ and
$\mathbf{\bar{B}} := \mathbf{B}-\mathbf{J}^T_F(\mathbf{J}_F\mathbf{M}^{-1}\mathbf{J}^T_F)^{-1}(\mathbf{J}_F\mathbf{M}^{-1}\mathbf{B})$~\cite{gao2019global}.

\subsubsection{Switching surface and impact dynamics}

When the swing foot strikes the DRS, an impact occurs, causing jumps in the generalized velocities $\mathbf{\dot{q}}$~\cite{grizzle2001asymptotically}.
The switching surface that determines the occurrence of the impact is given by
    \vspace{-0.05in}
\begin{equation}
\small
\begin{aligned}
	S_q:=\{(\mathbf{q,\dot{q}}): 
	z_{sw}(\mathbf{q})=0,
	\dot{z}_{sw}(\mathbf{q},\dot{\mathbf{q}})<0\}.
\end{aligned}
	\label{Eq:switch_sur_dp}
     \vspace{-0.05in}
\end{equation}

The velocity jump can be described by a reset map:
$\mathbf{\dot{q}}^+=\mathbf{R}_{{\dot{q}}}(\mathbf{q}^-)\dot{\mathbf{q}}^-$,
where $\mathbf{R}_{{\dot{q}}}$ is the reset map.
Different from the fixed-time switching of the ALIP model, the full-order switching surface is state-triggered, which is accurate in capturing the actual robot behavior.

\vspace{-0.05in}
\subsection{Input-output Linearizing Control}

The tracking error $\mathbf{y}$ is defined as $\mathbf{y}=\mathbf{h}_c(\mathbf{q})-\mathbf{h}_d(s)$.
With the following input-output linearizing control law~\cite{khalil1996noninear}
    \vspace{-0.05in}
\begin{equation}
\small
    \label{equ:feedback control law}
	\boldsymbol{\tau}_u=(\tfrac{\partial \mathbf{h}_c}{\partial \mathbf{q}} \mathbf{M}^{-1}\bar{\mathbf{B}})^{-1}[(\tfrac{\partial \mathbf{h}_c}{\partial \mathbf{q}} )\mathbf{M}^{-1}\bar{\mathbf{c}}
	+\mathbf{v}-\tfrac{\partial}{\partial \mathbf{q}}(\tfrac{\partial \mathbf{h}_c}{\partial \mathbf{q}}\mathbf{\dot{q})\dot{q}}+\tfrac{1}{T^2}\tfrac{\partial \mathbf{h}_d}{\partial s^2}],
     \vspace{-0.05in}
\end{equation}
the output function dynamics becomes $\dot{\mathbf{y}}=\mathbf{v}$, which is linear and time-invariant.
We can stabilize this linear system through proportional-derivative (PD) control, i.e., $\mathbf{v}=-\mathbf{K}_p\mathbf{y}-\mathbf{K}_d\mathbf{\dot{y}}$ with PD gains $\mathbf{K}_p$ and $\mathbf{K}_d$.

\noindent \textbf{Remark 4 (Effects of time-varying B\'ezier coefficients):}
The control law in Eq.~\eqref{equ:feedback control law} assumes the B\'ezier coefficients are constant within a continuous phase to completely cancel the nonlinearity of the continuous-phase dynamics in Eq.~\eqref{equ:full order dynamics with holo}.
Yet, in practice, to ensure walking robustness, the ALIP planner needs to update the desired footstep locations relatively frequently (e.g., 100 Hz) within a walking step, thus resulting in time-varying B\'ezier parameters.
As long as the B\'ezier parameters vary sufficiently slowly, the controller in Eq.~\eqref{equ:feedback control law} will still be effective thanks to its inherent robustness to model uncertainties.

\section{Simulation Validation}

\label{Section-Simulation}
This section reports the simulation validation results of the proposed approach for planar underactuated bipedal walking under different gait types and surface movements. 

\subsection{MATLAB Simulation Setup}


\subsubsection{Robot}
In MATLAB, a planar robot with point feet (Fig.~\ref{Fig:2d robot}) is simulated based on the hybrid full-order model given in Sec. V. 
The robot comprises one trunk link, two upper-leg links, and two lower-leg links, along with four motors.
The detailed mass distribution and geometric properties of this robot are given in Table~\ref{Table: robot info}.
The generalized coordinates of this robot are $\mathbf{q}= [p^b_x,p^b_y,\theta^b,q_1,q_2,q_3,q_4]$, where $q_1$,...,$q_4$ are the angles of the four revolute joints powered by the four motors.
As explained in Sec. IV-A, the robot has one degree of underactuation, and thus the dimension of the unactuated dynamics is two. 

\begin{table}[h]
\centering
\vspace{-0.05in}
\caption{\small{Mass distribution of the planar bipedal robot}} 
\small
\begin{tabular}{ p{2.5cm}|p{1.4cm}|p{1.7cm} }
\hline
\centering
Body component & \centering  Mass (kg) & ~~~{\centering  Length (m)} \\
\hline
\centering trunk & \centering 38 & ~~~0.63\\
\centering left/right thigh & \centering 0.3& {\centering ~~~0.4}\\
\centering left/right shank & \centering 0.3& ~~~0.4\\
\hline
\end{tabular}
\label{Table: robot info}
\end{table}

\subsubsection{DRS motions}

As this study focuses on horizontally and periodically moving surfaces, we use the following two swaying motions of the DRS in simulations: (M1) $x_S(t) = 0.03 \sin(\frac{2\pi}{0.4}t)$ m and (M2) $x_S(t) = 0.03 \sin(\frac{2\pi}{0.2}t)$ m.

\subsubsection{Desired gait types}
While this study addresses 2-D walking, two types of 2-D gaits are planned and tested, corresponding to the typical walking patterns in the sagittal and lateral planes of 3-D walking.
They are: (G1) robot walking along a single direction, with the two legs periodically passing each other, and (G2) robot stepping in place w.r.t. the DRS frame, without leg crossing.

\subsubsection{Simulated cases}
Two cases of DRS walking are simulated:
(A) the combination of the DRS motion (M1) and walking gait (G1)
and
(B) the combination of the DRS motion (M2) and walking gait (G2).

\subsubsection{Initial condition of full-order robot}

The initial state of the full-order model is obtained by solving a constraint problem that maps the state of the ALIP model to the full-order model while enforcing feasibility constraints.
Thus, the state of the full-order model can be mapped exactly to the corresponding ALIP model at the beginning of the simulation.
Let $\mathbf{q}_0=[{}^0p^{b}_x,{}^0p^{b}_y,{}^0\theta^{b},{}^0q_1,{}^0q_2,{}^0q_3,{}^0q_4]$ denote the obtained initial state.

\subsubsection{Implementation of higher-layer ALIP-based planner}
The fmincon command from MATLAB is used to solve the optimization problem in Eq.\eqref{equ:opt} for finding the two walking gaits (G1) and (G2).
The bounds on the step length and the ALIP state are chosen to represent the kinematic limits of the full-order robot, which are $u_{min}=-0.7 m$, $u_{max}=0.7 m$, $\mathbf{x}_{min} = [-0.7 m,-40 kgm^2/s]^T$, and $\mathbf{x}_{max} = [0.7 m,40 kgm^2/s]^T$.
We strengthen the stability constraint (C-3) in Eq.~\eqref{equ:opt} by setting $|\mu_i|<0.69$ instead of $|\mu_i|<1$. The smaller norm of the eigenvalues can lead to a faster convergence rate for the ALIP model, thus helping ensure the walking stability for the full-order robot despite the discrepancy between the ALIP and the actual robot dynamics.

For the simulation case (A), the user-defined parameter are chosen to be $H=0.81$ m, $m= 39.8$ kg, and $T = 0.4$ s.
For the simulation case (B), the parameters are the same as case (A) except $T=0.2$ s.
The optimization is solved in MATLAB within 10 seconds for both walking gaits.
The eigenvalues of case (A) are $\mu$= $-0.0231\pm0.0025i$. The eigenvalues of case (B) are $\mu= -0.3395\pm0.0001i$.

\subsubsection{Implementation of middle-layer full-body walking pattern generator}
The middle layer aims to provide the desired full-body trajectories $\mathbf{h}_d$ in Eq.\eqref{equ: hd}.
We choose $\phi_1=H$ to ensure the CoM of the full-order model is close to the ALIP model.
We choose $\phi_2= {}^0\theta^b$ so that the orientation of the robot trunk remains constant.
The B\'ezier coefficients of $\phi_4$ are chosen to be $[0, 0.075, 0.05, 0.045, 0.05, 0.075, 0]^T$.
We minimize the desired swing foot height to reduce the swing foot motion, leading to a smaller discrepancy between the full-order and the ALIP models.
The coefficients of $\phi_3$ are updated by the ALIP planner, which runs at $100$ Hz.

\subsubsection{Implementation of the lower-layer input-output linearizing controller}
The proposed full model based controller is implemented using Eq.~\eqref{equ:feedback control law}.
The matrices $\mathbf{M}$ and $\mathbf{c}$ are obtained using FROST~\cite{Hereid2017FROST}.
The control gains are chosen as
$\mathbf{K}_p=K_p\mathbf{I}$ (${K_p=2500}$) and $\mathbf{K}_d=K_d\mathbf{I}$ (${K_d=100}$) for both cases (A) and (B), where $\mathbf{I}$ is a $4\times 4$ identity matrix.
This choice of PD gains ensures the exponential convergence of the output function $\mathbf{y}$ within a continuous phase.

\subsection{Simulation Results}
\begin{figure}[t]
    \centering
    \includegraphics[width=0.9\linewidth]{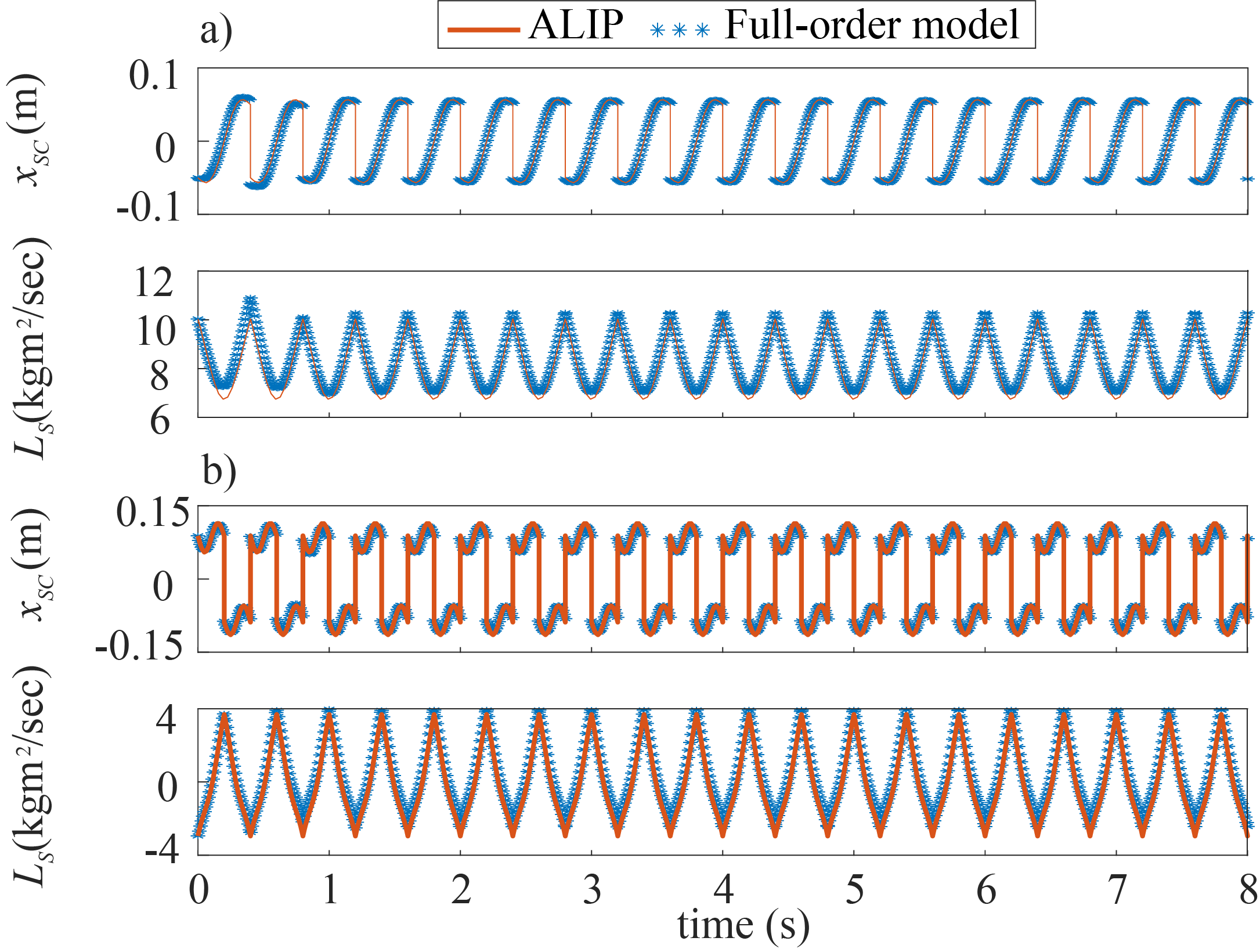}
    \vspace{-0.15in}
    \caption{State trajectories of the ALIP and the full-order robot during simulations of a) case (A) and b) case (B).
    The state trajectories of the full-order model stay close to those of the ALIP model, indicating a small model discrepancy.}
    \label{Fig:M1M2}
    \vspace{-0.2in}
\end{figure}

\begin{figure}[t]
    \centering
    \includegraphics[width=0.9\linewidth]{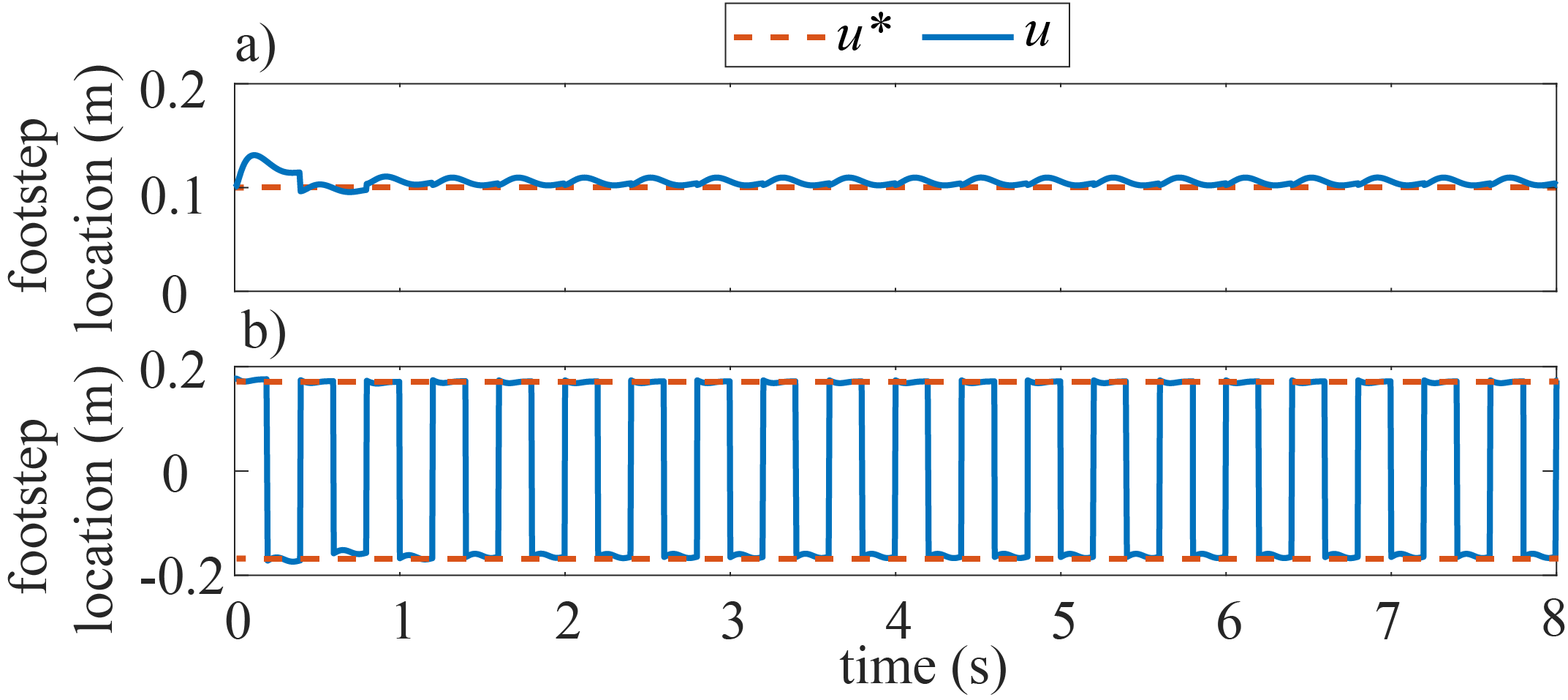}
    \vspace{-0.15in}
    \caption{Desired footstep location of the ALIP (i.e., $u^*$) and online-planned footstep location of the full-order robot (i.e., $u$) for a) case (A) and b) case (B).
    The small difference between $u^*$ and $u$ supports that the full-order robot tracks the desired ALIP footstep locations well.
    }
    \label{Fig:u}
    \vspace{-0.2in}
\end{figure}

The results of case (A) (Fig~\ref{Fig:M1M2} a)) and case (B) (Fig~\ref{Fig:M1M2} b)) indicate that the unactuated state variables of the full-order robot, i.e., the angular momentum about the contact point, $L_S$, and the relative CoM position, $x_{SC}$, remain bounded and close to the planned ALIP trajectories.
Figure~\ref{Fig:M1M2} confirms that our approach can ensure an underactuated, full-order robot model to walk stably on a DRS.

Figure~\ref{Fig:u} displays the desired and actual planned step lengths from case (A) and case (B). This result shows that the full-order model executes the desired footstep location well for both cases.

\begin{figure}[h]
    \centering
    \includegraphics[width=0.95\linewidth]{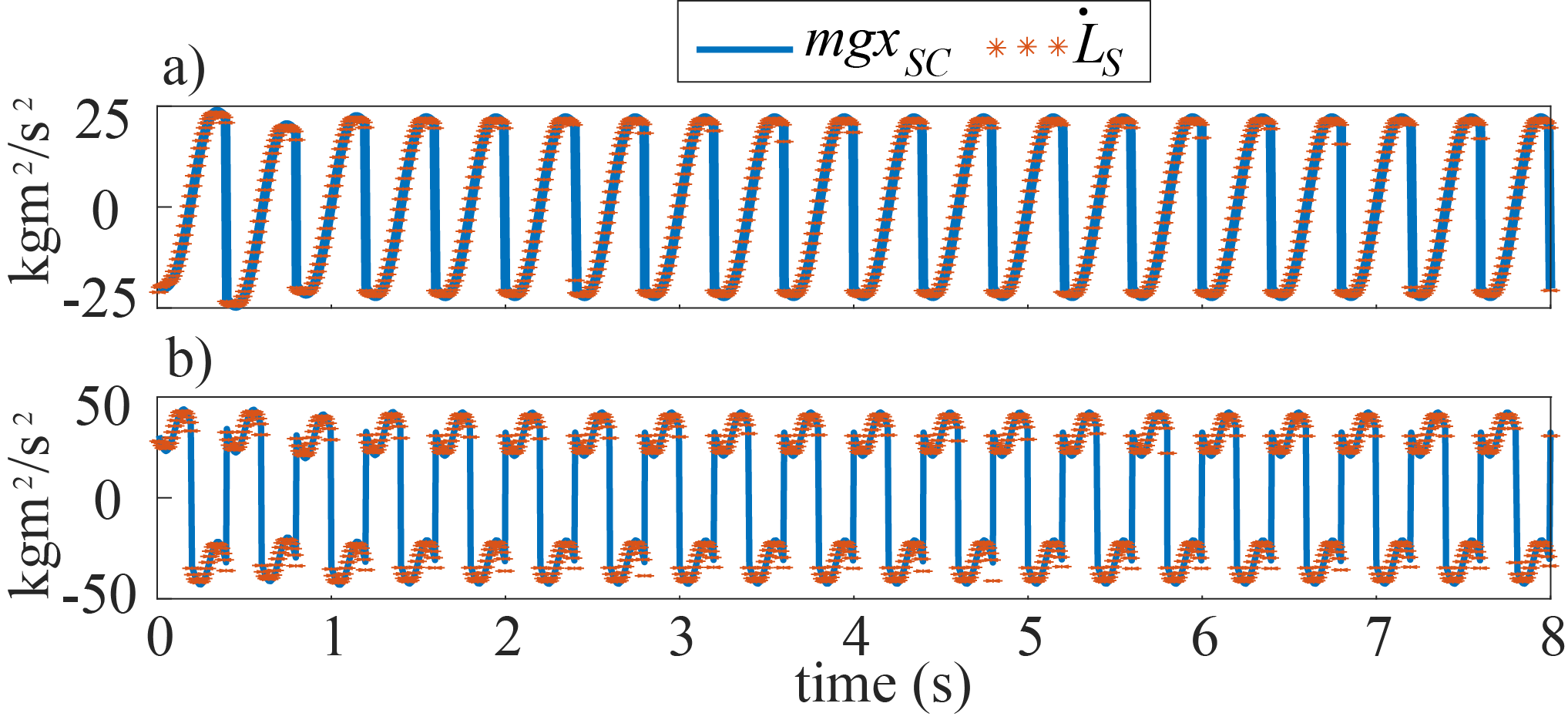}
    \vspace{-0.15in}
    \caption{Validation of $\dot{L}_S = mgx_{sc}$ for a) case (A) and b) case (B). 
    }
    \label{Fig:dl_x_sc.png}
    \vspace{-0.2in}
\end{figure}

\begin{figure}[h]
    \centering
    \includegraphics[width=1\linewidth]{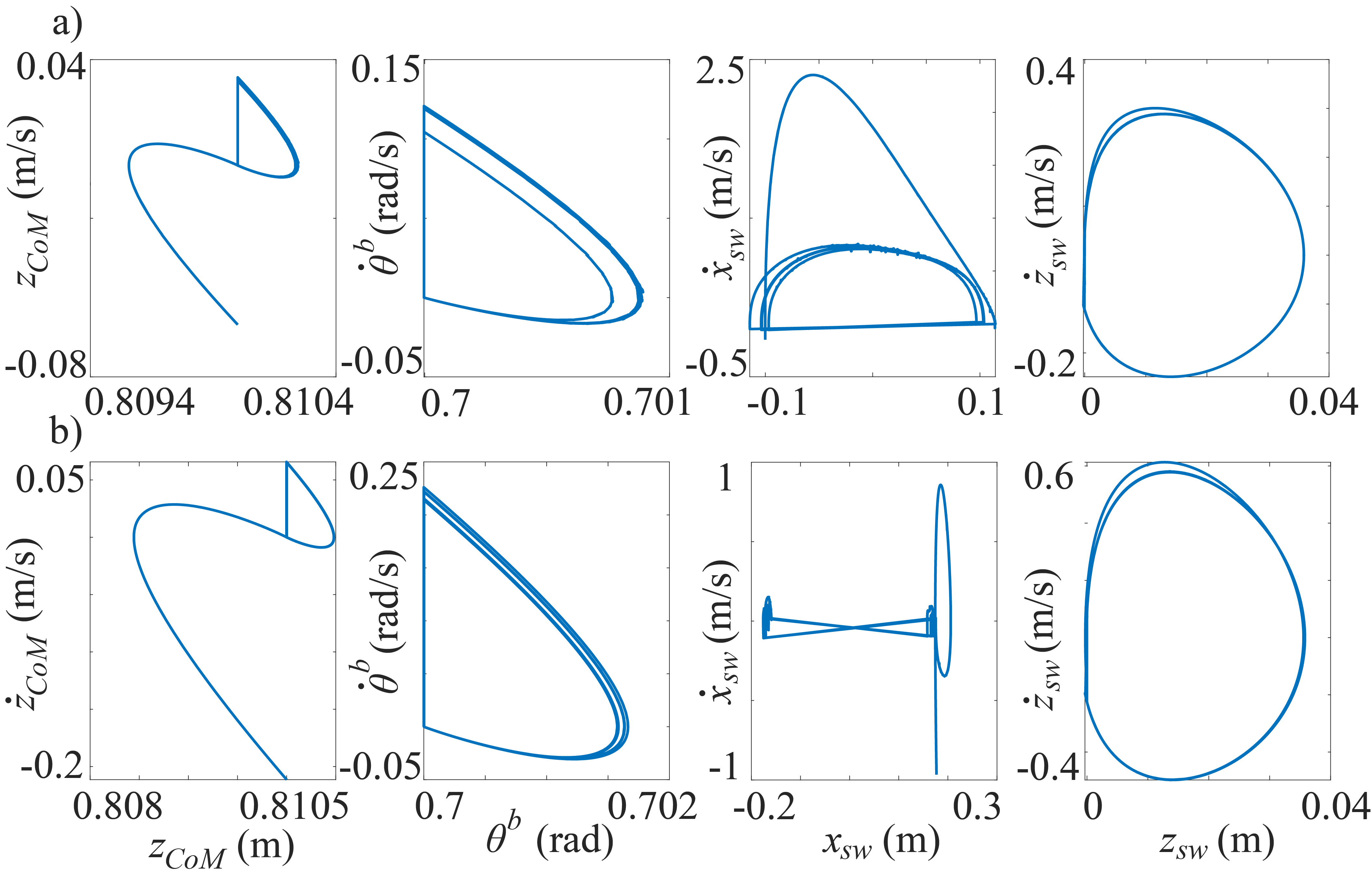}
    \vspace{-0.25in}
    \caption{Phase portraits of the directly controlled variables for a) case (A) and b) case (B), which indicate the robot's full-order motion is stable.}
    \label{Fig:phase_portrait}
    \vspace{-0.2in}
\end{figure}

There is a unique property of a point-feet robot walking without zero vertical CoM velocity: $\dot{L}_S = mgx_{sc}$, which is demonstrated in Fig.~\ref{Fig:dl_x_sc.png} 
Figure~\ref{Fig:dl_x_sc.png} shows that the vertical CoM velocity and angular momentum about the CoM are regulated well through the proposed middle-layer walking pattern generator and the lower-layer input-output linearizing controller. 
Figure~\ref{Fig:phase_portrait} shows the phase portrait of the controlled variables in Eq.~\eqref{equ:hc}. The plot indicates that the full-order robot is stable because no divergence is shown in the plot.

\section{Conclusions}
\label{section-conclusions}
This paper has proposed a control approach that enables underactuated bipedal robots to walk stably on a horizontally oscillating DRS.
An ALIP model for walking on a DRS was analytically derived with the time-varying surface movement and hybrid robot behaviors explicitly considered.
A discrete feedback control strategy was synthesized to stabilize the periodic solution of the hybrid ALIP model, whose continuous-phase dynamics is unstable and uncontrollable.
A hierarchical control approach was developed to stabilize the unactuated robot dynamics based on the hybrid ALIP model and its stabilizing footstep controller.
The proposed approach also included an input-output linearizing controller that ensures the exponential trajectory tracking for the directly commanded variables of the actual robot. 
Simulations of a planar full-order robot with point feet confirmed that the proposed framework stabilizes underactuated walking on a swaying DRS under different DRS motions and gait types.
Our future work will extend this framework to a 3-D underactuated bipedal robot and evaluate the performance on hardware.

\balance
\bibliography{Reference1}

\bibliographystyle{ieeetr}

\end{document}